\documentclass[11pt]{article}

\usepackage[margin=1in]{geometry}
\usepackage[T1]{fontenc}
\usepackage[utf8]{inputenc}
\usepackage{microtype}
\usepackage{amsmath,amssymb,amsfonts}
\usepackage{booktabs}
\usepackage{graphicx}
\usepackage{multirow}
\usepackage{xcolor}
\usepackage[numbers,sort&compress]{natbib}
\usepackage[colorlinks=true,linkcolor=blue!60!black,citecolor=blue!60!black,urlcolor=blue!60!black]{hyperref}

\newcommand{\C}{\mathbf{C}}
\newcommand{\q}{\mathbf{q}}
\newcommand{\kk}{\mathbf{k}}
\newcommand{\vv}{\mathbf{v}}
\newcommand{\ns}{\textsc{ns5}}
\newcommand{\solved}[2]{#1/#2}

\title{The Orthogonalized Read Is a Removable\\ Training Scaffold for Recurrent Memory}
\author{Keston Aquino-Michaels\\
  No Way Labs}
\date{July 2026}

\begin{document}
\maketitle

\begin{abstract}
Orthogonalizing the mLSTM memory matrix at read time with five differentiable
Newton--Schulz iterations improves noisy associative recall
\citep{tambde2026ortho}. We replicate this effect and investigate its mechanism.
Training on MAD noisy recall exhibits a long chance-level plateau followed by a
sharp increase in accuracy. The orthogonalized read improves conditioning during
this plateau and can be removed after escape. Ablations support three findings.
First, the benefit requires a self-consistent read and gradient: an exact
recursive least-squares read (the Mesa layer) yields a similar benefit, while
straight-through variants, delta-rule writes, frozen random keys, and Frobenius
normalization show no improvement over baseline. Second, across a learning-rate
$\times$ task-difficulty grid, orthogonalization multiplies escape hazard roughly
six-fold, with no detectable dependence on difficulty, and widens the range of
learning rates that produce successful runs. Third, adding orthogonalization at
inference leaves chance-level failures unresolved, while removing it gradually
after escape yields standard mLSTMs at near-perfect accuracy. Schedule changes
alone recover much of the reported gain. A batch-size $\times$ learning-rate
analysis separates the effects of per-step learning rate and gradient noise on
escape hazard (elasticities $+3.0$ and $-1.65$, respectively), linking the original
vocab-96 result to its large-batch training regime. Direct decoding of the memory
state recovers roughly half of the associations in behaviorally failed models,
indicating a readout-learning limitation despite substantial stored information.
These results show that fixed-budget recall benchmarks are sensitive to
trainability and provide a tractable setting for investigating abrupt behavioral
transitions through measurements of internal representations.
\end{abstract}

\section{Introduction}
\label{sec:intro}

Precise associative recall from context is the capability that separates softmax
attention from recurrent sequence models: the perplexity gap between the two
concentrates on recall-heavy tokens \citep{arora2023zoology,arora2024based}, and
the modern linear-recurrence program (mLSTM/xLSTM \citep{beck2024xlstm}, delta-rule
fast weights \citep{schlag2021linear,yang2024deltanet}, selective state spaces
\citep{gu2023mamba}, RWKV \citep{peng2023rwkv}) is largely an attempt to make a
fixed-size matrix memory support the precise recall provided by attention's
unbounded KV cache. Synthetic recall suites such as MAD \citep{poli2024mad}
and Zoology \citep{arora2023zoology} exist to select among these
architectures cheaply, at small scale, on the premise that the ranking transfers:
this is the explicit ``mechanistic design'' methodology \citep{poli2024mad}. A
model's solved-rate on the benchmark, at a fixed training budget, is read as its
recall capability.

We study read-time orthogonalization on MAD noisy recall and find that
fixed-budget solved-rate depends strongly on optimization dynamics. Training
exhibits a long chance-level plateau followed by a sharp escape. The fraction of
runs solved by a given budget therefore reflects the escape-time distribution,
with runs that remain on the plateau censored at the budget. Changes in schedule
and gradient noise alter this distribution, making architecture rankings
sensitive to the training configuration. This sensitivity raises a question for
mechanistic design: whether rankings obtained at small scale persist under the
schedules, budgets, and batch sizes used at larger scales. The task also offers a
tractable setting for studying abrupt behavioral transitions, since internal
representations can be probed while accuracy remains near chance
\citep{wei2022emergent,schaeffer2023mirage}.

The intervention we study is from \citet{tambde2026ortho} (henceforth ``the
original study''), which reports that a one-block mLSTM's performance on MAD noisy
associative recall improves dramatically when the memory matrix $\C_t$ is
orthogonalized at read time by five Newton--Schulz iterations
\citep{bjorck1971iterative,higham2008functions}, with gradients flowing through the
iteration. That study motivates the intervention by recall needs in attention-free settings
such as long-horizon model-based RL \citep{hafner2020dreamer}. Read-time
orthogonalization provides a focused test of how conditioning a superposed
matrix memory affects recall and training dynamics.

Our experiments identify the orthogonalized read as a removable training
scaffold: it improves conditioning during the plateau, and annealing it away
after escape preserves recall accuracy. We report seven findings:

\begin{enumerate}
\item \textbf{Replication} (\S\ref{sec:replication}). At the setup of
  \citet{tambde2026ortho} the effect reproduces: the orthogonalized read (\ns) roughly
  triples the fraction of
  seeds that solve noisy recall (9/16 vs.\ 3/16; pooled over two regimes 19/32
  vs.\ 7/32, Fisher $p=0.005$).
\item \textbf{The schedule confound} (\S\ref{sec:schedule}). Training is a
  chance-plateau $\to$ sharp-escape process, and solved-rate at a fixed budget is a
  censored measurement of escape time. Stretching the cosine period from 2{,}000 to
  4{,}000 steps lifts the baseline from 3/16 to 11/16 solved at step
  2{,}000, the same training compute (McNemar $p\approx0.008$). A constant
  learning rate alone reaches 8/16, which we cannot distinguish from \ns\
  (log-rank $p=0.41$; hazard ratio $1.5$, 95\% CI $[0.6, 4.3]$), at one sixth
  of \ns's per-step cost. Per unit wall-clock the schedule changes outperform
  \ns\ at this hardness: a wall-matched constant-LR baseline solves 14/16 in
  the time \ns\ needs for 9/16 (Fig.~\ref{fig:compute}); at the v96 frontier the
  ordering plausibly reverses (\S\ref{sec:basin}).
\item \textbf{Mechanism by elimination} (\S\ref{sec:mechanism}). The residual
  benefit requires self-consistent whitening of the read. An exact
  Sherman--Morrison/RLS read matches \ns; both straight-through halves (whitened
  forward with raw gradients, and vice versa) each solve 0/8 seeds; delta-rule
  erase-before-write keeps the memory well conditioned (effective rank 9.9
  vs.\ 2.6) without improving solved-rate; frozen random keys and read-path learning-rate
  boosts show no improvement; Frobenius normalization without the whitening iterations
  solves 1/16 seeds. Preserving memory rank alone is insufficient to
  improve escape.
\item \textbf{The scaffold property} (\S\ref{sec:scaffold}). Read-time
  orthogonalization rescues no failed baseline from chance at inference; 7/9 solved
  \ns\ models keep solving with the orthogonalized read removed (a 5--9 point accuracy loss);
  and annealing the read back to raw over the last 30\% of training recovers accuracy:
  12/16 seeds finish as numerically stock mLSTMs at 95--100\% accuracy.
  Triggering the anneal after escape also preserves success for a late-escaping
  seed: 13/16 solved, matching the un-annealed combination, with
  every solver finishing at $\alpha{=}0$, mean accuracy 99.6\% on the raw read.
\item \textbf{Where the scaffold matters} (\S\ref{sec:basin}). On a 144-run grid
  (three constant learning rates $\times$ three vocabulary sizes $\times$ two
  variants $\times$ eight seeds), the whitened read multiplies the escape hazard by
  $6.4\times$ (95\% CI $[3.5, 11.7]$; no detectable hardness interaction), rescuing
  the too-cold edge of the learning-rate corridor and suppressing destabilization at the too-hot
  edge. The baseline's hazard shrinks with hardness, so the realized gap
  grows even though the multiplier is flat.
\item \textbf{The escape law} (\S\ref{sec:batch}). Across a batch-size $\times$
  learning-rate grid, escape hazard rises with per-step heat (elasticity $3.0$,
  95\% CI $[2.0, 4.1]$) and falls with SGD gradient noise $\mathrm{lr}^2/B$
  (elasticity $-1.65$, $[-2.1, -1.2]$); their cancellation explains the
  learning-rate corridor, and batch size moves the corridor because it enters
  only the noise term. Reproducing the original study's headline regime (vocab 96,
  batch 64, its own schedule), the baseline improves from no solved runs to 1/8
  and the orthogonalized read to 5/8: a $+45$-point mean gain matching the
  original report, which we attribute to the escape-hazard multiplier acting on a
  baseline that large batch has lifted to the edge of escape. The batch-size dependence
  is consistent with the fitted effect of gradient
  noise on escape hazard.
\item \textbf{Storage/readout dissociation} (\S\ref{sec:decode}). Ridge-decoding
  the memory state directly: behaviorally failed models (${\sim}3\%$ accuracy)
  carry ${\sim}52\%$ linearly decodable, key-specific associations (wrong-key
  control ${\sim}5\%$). The plateau is a readout-learning failure over
  partially recoverable storage, consistent with the limited benefit of
  write-side conditioning.
\end{enumerate}

Two consequences extend beyond the intervention and are developed in
\S\ref{sec:discussion}. First, recall benchmarks used for architecture selection
partly measure trainability: when the underlying quantity is escape hazard, a
small-scale ranking reflects optimization ease as much as capability, and the
audit protocol used here (escape-time survival analysis, schedule and batch
robustness checks, storage/readout probes) applies directly to that practice.
Second, the system provides a directly instrumented example of an ``emergent''
threshold: capability-relevant structure is present and decodable while the
behavioral metric remains near chance, allowing the transition to be examined
through both optimization dynamics and internal representations.

\section{Setup}
\label{sec:setup}

\paragraph{Task.} MAD noisy in-context recall \citep{poli2024mad}: sequences of
key$\to$value token pairs with 80\% interleaved noise tokens drawn from a disjoint
distractor range; at each query position the model must emit the value previously
paired with the queried key. Vocabulary 80 (key/value ranges) unless stated,
sequence length 512, accuracy measured only at query positions. We vendor the MAD
generator; data are generated fresh each batch from a seeded stream.

\paragraph{Model.} One-block mLSTM \citep{beck2024xlstm}: embedding dim 94, 4
heads, pre/post LayerNorm, $\sim$78K parameters, the reference
implementation of \citet{tambde2026ortho} (verified against the author's code). Per
head, the mLSTM maintains a matrix memory
$\C_t=\sum_{u\le t} w_{t,u}\,\tfrac{\kk_u}{\sqrt{d_h}}\,\vv_u^{\!\top}$ with
gate-determined weights $w_{t,u}$, read as
$\mathbf{h}_t=\q_t^{\!\top}\C_t/\max(|\q_t^{\!\top}\mathbf{n}_t|,e^{-m_t})$.

\paragraph{The intervention.} \ns\ replaces the read by
$\q_t^{\!\top}\,\mathrm{NS}_5(\C_t)$, where $\mathrm{NS}_5$ is Frobenius
normalization followed by five Newton--Schulz iterations toward the nearest
orthogonal matrix, applied per position and trained through. Its per-position
materialization of $\C_t$ accounts for its computational cost: ${\sim}6\times$ baseline
wall time at our batch size, and $O(b\,s\,d_h^2)$ activation memory.

\paragraph{Variants.} Mechanism probes used throughout
(\texttt{memrec/models.py}): \textsc{ns}$k$ (with $k$ iterations; \textsc{ns0} =
normalization only), \textsc{ns5\_fwd}/\textsc{ns5\_bwd} (straight-through halves),
\textsc{delta} (erase-before-write, $\beta{=}1$), \textsc{rls} (Sherman--Morrison
whitened read $\q^{\!\top}(\sum\hat\kk\hat\kk^{\!\top}+\lambda I)^{-1}\C$, the
Mesa-layer read \citep{vonoswald2023mesa,vonoswald2025mesanet}),
\textsc{randk} (frozen random-orthogonal key projections), \textsc{pogo\_qk}
(q/k projections kept orthogonal by POGO \citep{javaloy2026pogo}), and read-path
learning-rate boosts. All variants share modules and training loop; state dicts
cross-load, which is what enables the swap evaluations of \S\ref{sec:scaffold}.

\paragraph{Training and metrics.} AdamW \citep{loshchilov2019adamw}, lr $3{\times}10^{-3}$,
batch 16, 2{,}000 steps with cosine decay ($T_{\max}$ = budget) unless stated,
the regime of \citet{tambde2026ortho}. Outcomes per seed are \emph{bimodal}: a run either
stays at chance (${\sim}3\%$) or escapes sharply to ${\ge}90\%$. We therefore emphasize
escape times and solved fractions, using mean accuracy
when comparing with the original report or describing solved checkpoints. The primary
metric is the escape-time distribution (first
evaluation with accuracy ${\ge}80\%$), analyzed with Kaplan--Meier estimates
\citep{kaplan1958nonparametric}, log-rank tests \citep{mantel1966evaluation} with
runs censored at budget, and discrete-time hazard models \citep{allison1982discrete};
solved-fraction tables (Fisher exact, McNemar for seed-paired comparisons) are
reported alongside. Batch 16 (vs.\ 64 in the original study) keeps every regime a
fair A/B under identical compute; \S\ref{sec:batch} treats batch size explicitly.
Two contrasts are designated primary and confirmatory: \ns\ vs.\ baseline escape
hazard at the original schedule, and the seed-paired schedule-stretch flip
(McNemar). All other comparisons are exploratory; we report their unadjusted
$p$-values as graded evidence and draw no individual confirmatory claim from
them, and we accompany null results with effect-size confidence intervals to convey the
uncertainty in those estimates.

\begin{figure}[t]
\centering
\includegraphics[width=0.78\linewidth]{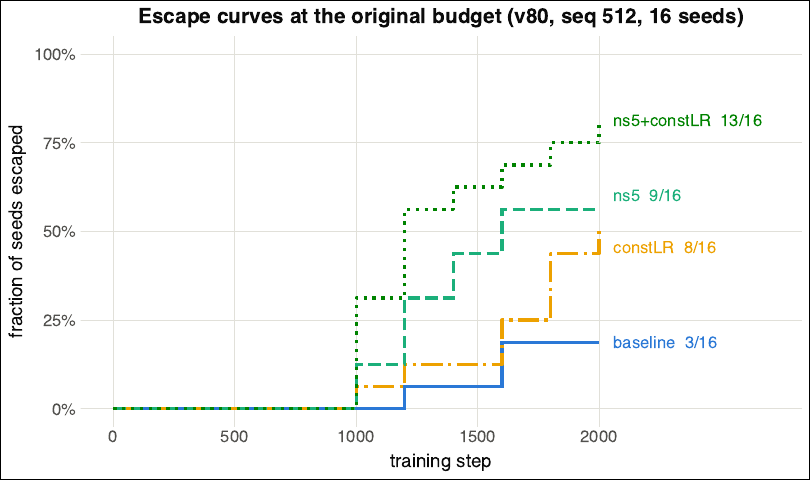}
\caption{Escape curves at the original budget of \citet{tambde2026ortho} (v80,
seq 512, cosine $T_{\max}{=}2000$ except where noted; 16 seeds each). Curves are
Kaplan--Meier estimates of the fraction of seeds escaped by each step, one per
training configuration. Claims: \S\ref{sec:replication} and
\S\ref{sec:schedule}.}
\label{fig:survival}
\end{figure}

\section{Replication}
\label{sec:replication}

In the regime of \citet{tambde2026ortho} our reimplementation reproduces the
direction and rough size of the published effect (Fig.~\ref{fig:survival}): \ns\ solves
\solved{9}{16} seeds
vs.\ the baseline's \solved{3}{16} at v80/s512, \solved{10}{16} vs.\ \solved{4}{16}
at v80/s1024 (pooled 19/32 vs.\ 7/32, Fisher $p=0.005$; log-rank on escape times
$p=0.020$). The author's code reproduces the same direction at batch 16 at
vocab~80. The original headline vocab-96 regime does not reproduce at batch 16 under
its own schedule: every variant collapses to chance; the original needed batch 64
on an A100. We return to both facts (\S\ref{sec:basin}, \S\ref{sec:batch}): these
dependencies motivate the analysis of optimization dynamics below.

\section{The schedule confound: solved-rate is censored escape time}
\label{sec:schedule}

\begin{figure}[t]
\centering
\includegraphics[width=0.78\linewidth]{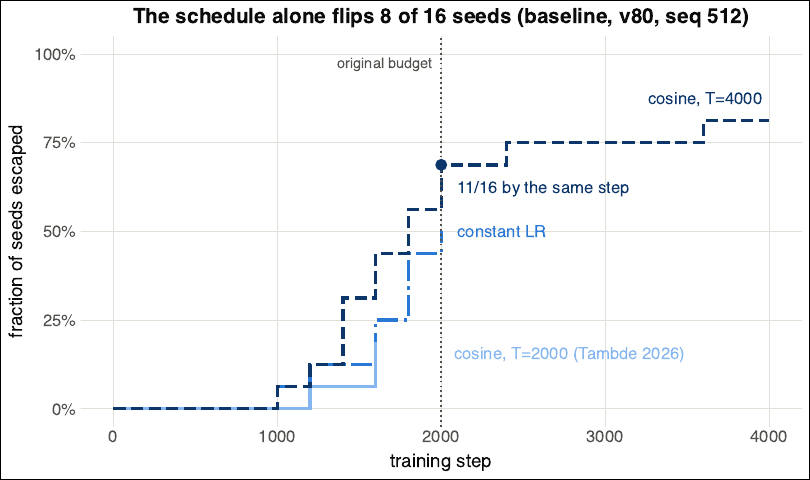}
\caption{Baseline mLSTM escape curves under schedule variation (v80/s512; same
16 seeds per curve). Curves are Kaplan--Meier estimates as in
Fig.~\ref{fig:survival}, one per schedule. Claims: \S\ref{sec:schedule}.}
\label{fig:schedule}
\end{figure}

Training on this task sits at chance for hundreds to thousands of steps, then
transitions sharply (Fig.~\ref{fig:anneal} shows example traces). The shape is
familiar from grokking \citep{power2022grokking}, although the training regimes differ:
classic grokking is delayed generalization after the training set has been
memorized, whereas query accuracy on this task remains near chance before escape; the
escape-dynamics analysis carries over
(\S\ref{sec:batch}). A fixed step
budget with a decaying learning rate therefore censors the escape process
twice: runs that would escape later are cut off by the budget, and the
decay reduces the learning rate during the interval in which escape may occur.

\begin{figure}[t]
\centering
\includegraphics[width=0.78\linewidth]{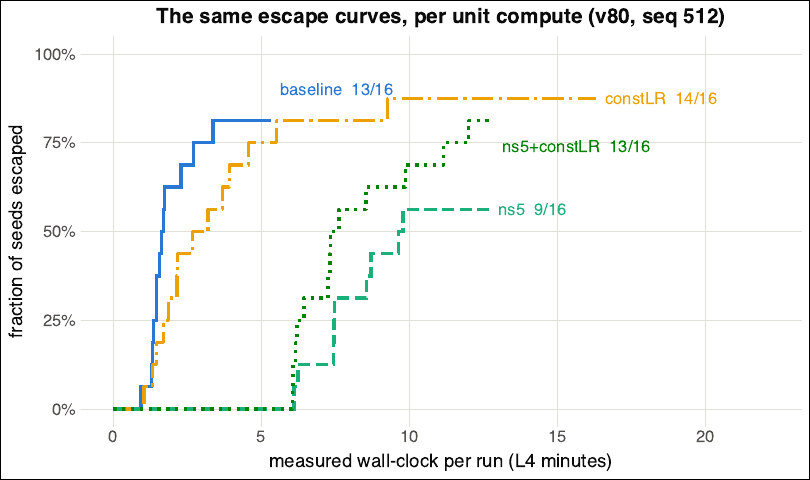}
\caption{Escape fraction against each run's measured wall-clock on identical L4
GPUs. Each family appears under its best-performing tested schedules: baseline
under cosine $T_{\max}{=}4000$ and under constant LR with a wall-matched
12{,}000-step budget; \ns\ under its published schedule and with the constant-LR
schedule. Claims and the pairing rationale: \S\ref{sec:schedule}.}
\label{fig:compute}
\end{figure}

Fig.~\ref{fig:schedule} isolates the effect. With the cosine period stretched from
$T_{\max}{=}2{,}000$ to $4{,}000$ (same seeds), the baseline solves
\textbf{11/16 by step 2{,}000}, the step count at which the original schedule
yields 3/16, and 13/16 by step 4{,}000. Eight seeds flip failed$\to$solved from
the schedule alone (McNemar $p\approx0.008$). The same pattern holds at s1024
(4/16 $\to$ 10/16 $\to$ 15/16). Deleting the decay entirely (constant LR, stop at
2{,}000) yields 8/16 with five more seeds mid-escape (57--70\%) at
cutoff. We cannot distinguish this from \ns\ at the original budget: log-rank
$p=0.41$, and a two-group discrete-time fit puts the \ns-vs-constant-LR hazard
ratio at $1.5$ with 95\% CI $[0.6, 4.3]$. Sixteen seeds per arm leave effects up
to ${\sim}4\times$ in either direction compatible, so the comparison leaves substantial
uncertainty about the relative hazards.
The observed ordering under matched wall-clock budgets is clearer at this
task difficulty. \ns's per-position Newton--Schulz graph
costs ${\sim}6\times$ baseline wall time per step; Fig.~\ref{fig:compute}
therefore compares each family under its best-performing tested schedules to compare
solved fractions under a fixed
wall-clock budget. Given
\ns's ${\sim}12$-minute budget (12{,}000 constant-LR steps), the plain baseline
solves \textbf{14/16}, above \ns's 9/16 and \ns{}+constLR's 13/16 at the same
wall cost. At the
v96 frontier, where no tested budget makes the baseline reliable, the
per-compute ordering plausibly reverses (\S\ref{sec:basin}).

\ns\ retains an advantage at matched schedules: 16/16 at
$T_{\max}{=}4{,}000$ including all three seeds unsolved by the baseline, with
systematically earlier escapes (median step 1{,}200 vs.\ 1{,}600; log-rank
$p=0.031$). We next examine the mechanism of this advantage (\S\ref{sec:mechanism}),
its persistence after removal of the intervention (\S\ref{sec:scaffold}), and
its dependence on task difficulty (\S\ref{sec:basin}).

\section{Mechanism by elimination: self-consistent whitening of the read}
\label{sec:mechanism}

During the plateau, the memory matrix is noise-dominated: 80\% of writes are
distractor tokens, and $\C$'s effective rank collapses from ${\sim}11$ at
initialization to ${\sim}2.5$ within 200 steps. The matched-filter read
$\q^{\!\top}\C$ over such a matrix defines an ill-conditioned learning problem for
the read path (q/k projections): gradient coherence between independent batches,
measured on the q/k projections during the plateau, is 0.20 for the baseline
vs.\ 0.42 for \ns: the whitened read doubles read-path gradient
signal-to-noise while full-model coherence is identical (0.44 vs.\ 0.42).

Table~\ref{tab:elimination} compares candidate mechanisms at the original
schedule.

\begin{table}[t]
\centering
\small
\begin{tabular}{llccc}
\toprule
variant & what it changes & solved & \shortstack{q/k grad-\\coherence} & \shortstack{$\C$ eff.\\rank} \\
\midrule
baseline & -- & 3/16 & 0.20 & 2.6 \\
\textsc{delta} (erase-before-write) & state conditioning & 2/16 & 0.21 & \textbf{9.9} \\
\textsc{randk} (frozen random keys) & measurement ensemble & 3/16 & -- & -- \\
q/k LR $\times4$ & optimizer pressure on read path & 4/16 & -- & -- \\
\textsc{ns5\_fwd} (NS fwd, raw grads) & forward only & \textbf{0/8} & -- & -- \\
\textsc{ns5\_bwd} (raw fwd, NS grads) & backward only & \textbf{0/8} & -- & -- \\
\textsc{ns0} (normalize, 0 iters) & scale only, no whitening & 1/16 & -- & -- \\
\textsc{ns1} (one NS iteration) & partial whitening & \textbf{0/16} & 0.27 & 3.3 \\
\textsc{ns3} (three NS iterations) & most of the whitening & 6/16 & 0.41 & 3.8 \\
\ns\ \citep{tambde2026ortho} & self-consistent spectral whitening & \textbf{9/16} & \textbf{0.42} & 4.2 \\
\textsc{rls} (Sherman--Morrison read) & self-consistent exact whitening & \textbf{5/8} & 0.28 & 4.3 \\
baseline, constant LR & schedule only & \textbf{8/16} & \multicolumn{1}{c}{0.45 / 0.20$^{\dagger}$} & -- \\
\ns\ + constant LR & read and schedule & \textbf{13/16} & -- & -- \\
\bottomrule
\end{tabular}
\caption{The elimination table. All cells: v80/s512, cosine $T_{\max}{=}2000$
unless noted, 16 seeds (8 for \textsc{rls} and straight-through cells). Plateau
metrics averaged over steps 200--600, before any escape.
$^{\dagger}$escapers/non-escapers at step 600 within constant-LR baseline runs.}
\label{tab:elimination}
\end{table}

\begin{figure}[t]
\centering
\includegraphics[width=\linewidth]{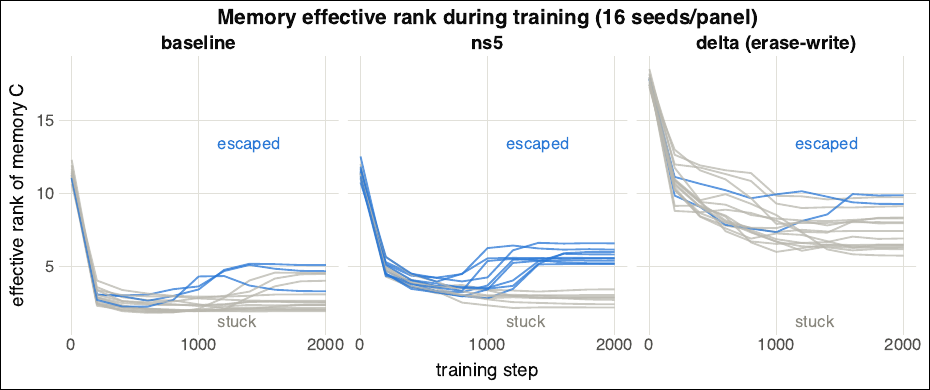}
\caption{Effective rank of the raw memory $\C$ (final position, first validation
batch) over training, for baseline, \ns, and \textsc{delta} runs; one line per
seed, colored by outcome. Claims: \S\ref{sec:mechanism}.}
\label{fig:rank}
\end{figure}

The ablations support five observations:

\textbf{Preserving state rank is insufficient.} The delta-rule write
preserves the effective rank of $\C$ (effective rank 9.9 vs.\ the baseline's 2.6,
Fig.~\ref{fig:rank}) and solves 2/16 seeds. This result indicates that preventing rank collapse
alone is insufficient to resolve the plateau.

\textbf{Key-Gram rank is similar across outcomes.} The key-Gram effective rank is
indistinguishable across variants and outcomes (${\approx}5.5$--$5.8$); frozen
random-orthogonal keys yield the same solved fraction as baseline. These
observations provide no evidence that degradation of the key measurement matrix
explains recall failure.

\begin{figure}[t]
\centering
\includegraphics[width=0.62\linewidth]{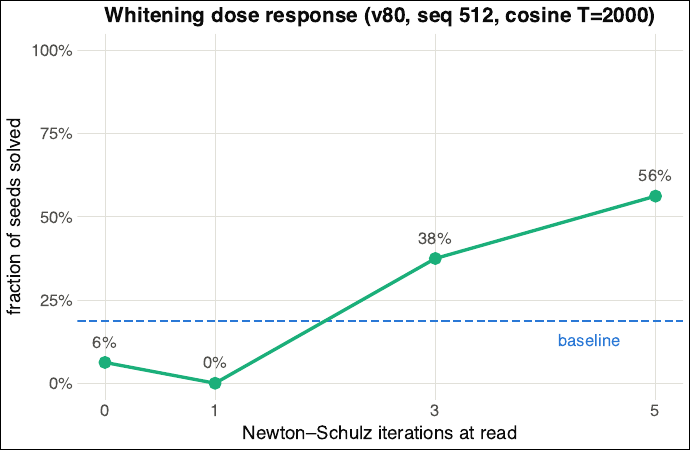}
\caption{Whitening dose response: fraction of seeds solved vs.\ Newton--Schulz
iteration count at the read (16 seeds per point, original schedule; dashed line =
baseline mLSTM at 3/16). Claims: \S\ref{sec:mechanism}.}
\label{fig:dose}
\end{figure}

\textbf{The benefit depends on whitening depth (Fig.~\ref{fig:dose}).}
The dose response is monotone above $k{=}1$ and thresholded. \textsc{ns0} (Frobenius
normalization, zero iterations) solves 1/16 seeds. A single Newton--Schulz
iteration yields a lower solved fraction than baseline (0/16 vs.\ 3/16), with plateau
read-path coherence (0.27) only slightly above the baseline's 0.20. The samples
are too small to establish the dip (Fisher $p=0.23$; log-rank $p=0.07$), but its
direction matches the straight-through cells: a partially whitened read, like an
inconsistently whitened one, may impair optimization relative to the raw read. Three iterations
recover nearly all of the coherence (0.41 vs.\ \ns's 0.42) and most of the
benefit (6/16 vs.\ 9/16) at 70\% of \ns's wall time (mean 8.5 vs.\ 12.3 minutes
per run, baseline 2.0); the escape statistics of \textsc{ns3} sit between
baseline and \ns\ and do not separate from either at these sample sizes. The
lowest tested iteration count with an observed benefit is $k{=}3$.

\textbf{Forward and backward consistency is required in these tests.} Both straight-through cells are at or below
baseline (0/16 pooled; decisively worse than \ns, Fisher $p\approx8\times10^{-4}$).
Both variants use gradients that differ from the derivative of the forward
read. Their failure supports the importance of differentiating through the
whitened read, which we refer to as self-consistent whitening.

\textbf{An exact least-squares read yields a similar benefit.} The recurrently
computed regularized least-squares read in \textsc{rls} is the
Mesa layer of \citet{vonoswald2023mesa}, scaled up in MesaNet
\citep{vonoswald2025mesanet}, and the space it spans (raw, delta-rule, and
least-squares reads as choices of regressor and optimizer) is the one
formalized by test-time regression \citep{wang2025testtime}, of which
Newton--Schulz is a spectral approximation and the delta rule an
online-gradient one. It solves 5/8 seeds with earlier escapes (log-rank
vs.\ baseline $p=0.022$, vs.\ \ns\ $p=0.92$) in a recurrent $O(d_h^2)$/step
form: a recurrent least-squares implementation therefore also supports improved
trainability through a self-consistent conditioned read.

Finally, read-path gradient coherence is associated with subsequent escape:
pre-escape coherence discriminates future escapers within every route measured
(within constant-LR baseline runs: escapers 0.45 vs.\ non-escapers 0.20 at step
600), but boosting the read-path learning rate
$4\times$ yields only 4/16 solved seeds, and no plateau probe at a fixed early step predicts
when a seed escapes (all $p>0.2$). Elevated coherence is the shared
signature of runs approaching escape; the interventions that improve escape
are self-consistent read whitening and learning-rate schedule changes.

\section{A removable scaffold}
\label{sec:scaffold}

If the whitened read works by re-conditioning the learning problem, it should be
unnecessary once learning is done. Three tests confirm this.

\paragraph{Swap evaluations.} Re-evaluating trained checkpoints under the other
read: adding $\mathrm{NS}_5$ to failed baselines at inference rescues
\textbf{none from chance} (12/12 at-chance failures stay at chance; the single
borderline seed, at 78\%, increases to 82\%), showing that read-time orthogonalization
alone is insufficient to recover
recall in these failed checkpoints. Conversely \textbf{7/9} solved \ns\ models keep solving with the
orthogonalized read removed (mean 97\%$\to$88\%; a consistent 5--9 point accuracy loss), and
$\mathrm{NS}_5$ on solved baselines is a $+1.6$ point accuracy improvement.

\begin{figure}[t]
\centering
\includegraphics[width=0.78\linewidth]{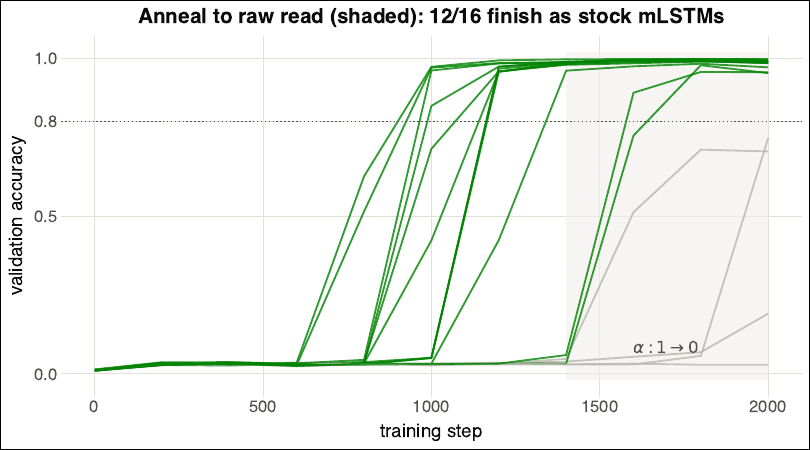}
\caption{Accuracy traces under the fixed-window scaffold anneal (window shaded):
the orthogonalized read is interpolated back to the raw read over steps
1400--2000; one line per seed, colored by outcome. Claims: \S\ref{sec:scaffold}.}
\label{fig:anneal}
\end{figure}

\paragraph{Scaffold anneal.} Interpolating the read
$\alpha\,\mathrm{NS}_5(\C)+(1-\alpha)\,\C$ from $\alpha{=}1$ to $0$ over the final
30\% of training (with constant LR) recovers near-perfect accuracy: \textbf{12/16 seeds
finish at 95.4--99.8\% (mean 98.4\%) with $\alpha{=}0$}; the final model is
numerically a standard mLSTM with zero inference overhead (Fig.~\ref{fig:anneal}).
The un-annealed combination solves 13/16; the additional failure under the
fixed window occurs in a seed that escapes during annealing.

\paragraph{Escape-gated anneal.} Triggering the anneal on the first evaluation
$\ge$80\% (200-step dwell, 500-step ramp, budget extended when needed so late
escapers finish their ramp) removes the fixed-window failure mode. The gate fires
only after a run has escaped within the nominal budget. Additional steps
complete scaffold removal and allow post-escape accuracy to improve; the
reported escape count is determined within the nominal budget.
Sixteen seeds under the gate (constant LR, 200-step dwell, 500-step ramp):
\textbf{13/16 solved, all thirteen at $\alpha{=}0$, mean 99.6\%} (min 99.3\%) on
the raw read: the un-annealed combination's solve rate, now with zero
inference overhead and near-perfect accuracy. Trigger steps ranged from 1{,}000 to
2{,}000; nine runs extended their budget (at most $+900$ steps, all post-escape)
to complete the ramp. The three failures never escaped and are consistent with the
failure mode of the un-annealed combination.

The procedure also succeeds at higher task difficulty. At the v96 frontier (constant LR,
4{,}000-step budget, the regime where the scaffold's advantage is largest), the
escape-gated anneal solves \textbf{7/8, all seven finishing at $\alpha{=}0$ on
the raw read} at 97.6--100\% (mean 98.6\%). The full pipeline (scaffold,
escape, removal) therefore delivers a numerically stock mLSTM at frontier
accuracy in the v96 regime.

\section{Where the scaffold matters: the basin map}
\label{sec:basin}

\begin{figure}[t]
\centering
\includegraphics[width=\linewidth]{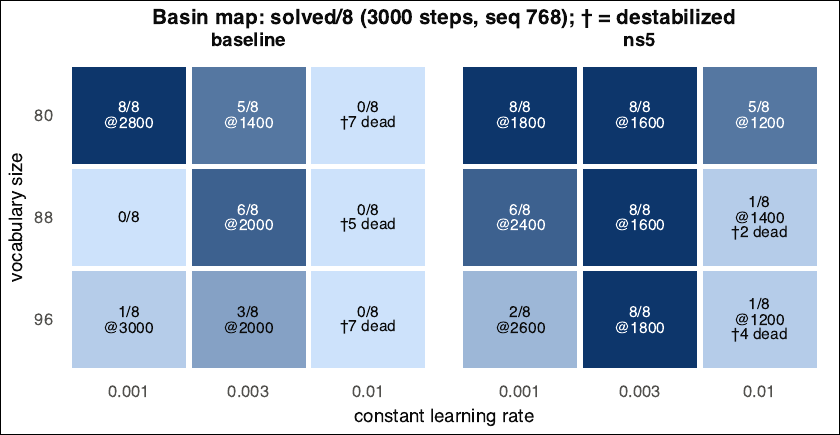}
\caption{Basin map: solved seeds out of 8 per cell (constant LR $\times$
vocabulary size, seq 768, 3{,}000 steps), with median escape step and
destabilization counts ($\dagger$ = accuracy collapses below chance, a
competing risk distinct from plateauing). Claims: \S\ref{sec:basin}.}
\label{fig:basin}
\end{figure}

To measure how the benefit varies with learning rate and task difficulty, we ran
constant LR
$\{10^{-3},3{\times}10^{-3},10^{-2}\}$ $\times$ vocab $\{80,88,96\}$ at seq 768
$\times$ \{baseline, \ns\} $\times$ 8 seeds $\times$ 3{,}000 steps
(Fig.~\ref{fig:basin}). The grid supports four observations:

\textbf{The baseline's corridor narrows with hardness.} v80 solves at two of three
learning rates; v88 at exactly one ($10^{-3}$ drops to 0/8); v96 at none fully
(best 3/8). Higher task difficulty reduces the range of learning rates that yield
consistent success within the tested budget.

\textbf{Whitening widens the corridor in both directions.} Cold edge: v88 at
$10^{-3}$ goes 0/8 $\to$ 6/8 (log-rank $p=0.0025$). Hot edge: baselines
destabilize at $10^{-2}$ (19/24 collapse below chance across regimes) while
\ns\ at v80 has zero collapses and solves 5/8, though this stability protection
fades with hardness (destabilized: 0/8 $\to$ 2/8 $\to$ 4/8). Center: \ns\ is 24/24 at
$3{\times}10^{-3}$ across all vocabulary sizes vs.\ the baseline's 14/24.

\textbf{A uniform multiplier under a shrinking base rate.} A discrete-time hazard
model (1{,}633 run-intervals, 70 escapes; destabilized runs censored at collapse as
a competing risk) gives \ns\ a $6.4\times$ escape-hazard multiplier (95\% CI
$[3.5, 11.7]$). The fitted \ns$\times$hardness interactions are small relative to their uncertainty
(\ns$\times$v88: HR $1.03$, 95\% CI $[0.29, 3.7]$; \ns$\times$v96: $0.74$,
$[0.18, 3.1]$), consistent with a hardness-independent multiplier, though with
70 escape events the intervals leave factor-of-three modulation in either
direction compatible. The baseline hazard itself falls with hardness (v96:
$0.14\times$, $z=-5.5$). A flat multiplier on a shrinking base
produces the growing realized advantage, including the headline vocab-96
regime of \citet{tambde2026ortho}. (An earlier fit on cosine-schedule data suggested the
multiplier itself grew with hardness; the constant-LR grid shows that
apparent interaction was attributable to schedule-dependent censoring.)

\textbf{The learning-rate effect is non-monotone.} Relative to $10^{-3}$, hazard at
$3{\times}10^{-3}$ is $4.9\times$ but at $10^{-2}$ is $0.23\times$, plus
the competing destabilization risk. A pure cumulative-heat law
($\int \mathrm{lr}\,dt$) fails: v80 escapes at $10^{-3}$ on less total heat
than $3{\times}10^{-3}$ escapes use, while the cold column at v88/v96 is still at
chance or mid-climb at budget (remaining censored at the longer budget). A candidate
explanatory variable is SGD noise ($\propto \mathrm{lr}^2/B$), which motivates the
batch axis.

At matched constant-LR schedules the frontier regime shows the practical
consequence: at v96/s768 with 4{,}000 steps, the baseline manages 1/4 solved (plus
two mid-escapes) while \ns\ solves 4/4: the original headline regime, reached at
roughly $1/8$ of its original compute. At this difficulty, the per-compute comparison may favor the
whitened read (4/4 at ${\sim}6\times$ per-step cost, against a baseline no tested
budget makes reliable), though we have not run a wall-matched baseline arm at this
regime. The realized advantage grows with hardness as the
uniform-multiplier law predicts. Combining the interventions improves the observed
solved fraction: \ns{}+constant LR at the original 2{,}000-step budget yields the
highest solved fraction among
step-matched configurations (13/16; log-rank vs.\ constant LR alone
$p=0.017$). Per unit compute the ranking differs (\S\ref{sec:schedule},
Fig.~\ref{fig:compute}): a wall-matched constant-LR baseline reaches 14/16, since
each \ns\ step costs ${\sim}6\times$ a baseline step.

\section{The batch axis: heat and noise in the escape law}
\label{sec:batch}

\begin{figure}[t]
\centering
\includegraphics[width=0.6\linewidth]{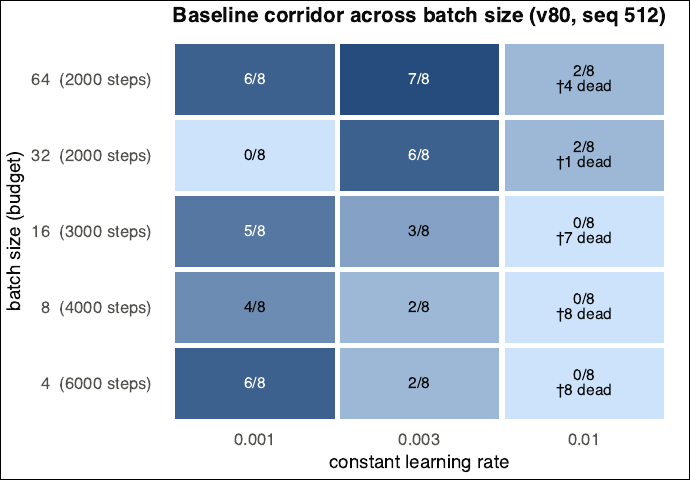}
\caption{Baseline solved fractions across batch size $\times$ constant LR
(v80/s512, 8 seeds per cell). Budgets scale inversely with batch (printed per
row) to allow more optimization steps in small-batch cells; the hazard model of
\S\ref{sec:batch} accounts for the differing budgets through censoring. Claims:
\S\ref{sec:batch}.}
\label{fig:batch}
\end{figure}

The basin map varied heat; batch size varies gradient noise at fixed heat. We ran
the baseline at batch $\{4,8,16,32,64\}$ $\times$ constant LR
$\{10^{-3},3{\times}10^{-3},10^{-2}\}$ $\times$ 8 seeds (v80/s512;
Fig.~\ref{fig:batch}), an \ns\ arm at batch $\{4,8,16\}$ at the reference LR, and
one replication arm at the original study's own setting.

\textbf{The corridor moves with batch size.} At batch $\le$16 the workable
learning rate is $10^{-3}$ and $10^{-2}$ causes destabilization (23/24 destabilized); at
batch 32--64 the corridor's center shifts to $3{\times}10^{-3}$ (6/8 and 7/8) and
the higher learning rate becomes more stable. The main study's batch of 16 (chosen so
\ns\ fits a 24\,GB GPU) therefore sits in a noisier, narrower-corridor regime
than the original study's batch 64, with consequences for replication, as shown below.

\textbf{The escape law separates heat from noise.} A discrete-time hazard fit
over the 120 baseline runs (1{,}645 run-intervals, 45 escapes; runs censored at
collapse or their own budget) with terms $\log t$, $\log \mathrm{lr}$, and
$\log B$ rejects both simple laws: a pure cumulative-heat law predicts no batch
effect ($\log B$ coefficient: $1.65$, $z=6.8$), and a tokens-processed law
predicts equal coefficients on $\log B$ and $\log \mathrm{lr}$ (the latter is
$-0.26$, $z=-1.4$). Refitting the same likelihood in the (heat, noise) basis,
with $\log \mathrm{lr}$ at fixed noise and $\log(\mathrm{lr}^2/B)$ (the standard
SGD gradient-noise scale), gives the following estimates: escape hazard rises
with per-step heat with elasticity $3.0$ (95\% CI $[2.0, 4.1]$) and falls with
gradient noise with elasticity $-1.65$ ($[-2.1, -1.2]$). The near-null raw
learning-rate coefficient is their cancellation ($3.04 - 2\times1.65 = -0.26$):
the heat term scales with learning rate, while the noise term scales with its
square. This
is the corridor of \S\ref{sec:basin}: its cold edge is
heat starvation, its hot edge is noise plus the destabilization risk, and
batch size relocates it because $B$ enters the noise term alone. A related theoretical
analysis uses the same variables:
\citet{ersoy2026metastable} model grokking as noise-driven escape from a
metastable phase and derive escape-time scaling in $B/\eta$ for linear networks;
our fit is the empirical decomposition on a recall task, under a different model and task setting.

\textbf{The whitened read improves escape at smaller batch sizes.} At the
reference LR, \ns\ vs.\ baseline by batch: 6/8 vs.\ 2/8 at batch 4 (log-rank
$p=0.049$), 7/8 vs.\ 2/8 at batch 8 ($p=0.006$), 13/16 vs.\ 8/16 at batch 16
($p=0.017$). Orthogonalization improves solved fractions at smaller batch sizes,
consistent with the uniform-multiplier interpretation of \S\ref{sec:basin}.

\textbf{Batch size affects replication of the vocab-96 result.} We compare
baseline and orthogonalized reads under the original study's headline regime. Under its
own cosine schedule at v96/s768, batch 64 takes the baseline from
chance-level performance in all runs (batch 16, \S\ref{sec:replication}) to 1/8 solved, 2/8
mid-escape, mean 22\%. Adding the orthogonalized read at that same setting gives
5/8 solved, mean 67\%: a $+45$-point mean gain, matching the original study's
reported $+40$-point delta at vocab 96. The observed gain is consistent with the
${\sim}6\times$ escape-hazard multiplier (\S\ref{sec:basin}) acting in a regime
where the larger batch enables some baseline runs to begin escaping. Together
with the batch-grid analysis, this result supports a gradient-noise explanation
for the dependence on batch size.

\section{Inside a failed model's memory}
\label{sec:decode}

The limited benefit of write-side interventions (Table~\ref{tab:elimination})
motivates a direct test of stored information. We used ridge regression to decode
the value
token from the feature $\kk_u^{\!\top}\C_t$ (the model's own stored key
at the pair's write position, slicing its own memory at query time),
fit on a train split of held-out validation sequences and scored on unseen
sequences, with a wrong-key control (the same pipeline keyed by a different stored
pair from the same sequence).

\begin{table}[h]
\centering
\small
\begin{tabular}{lccc}
\toprule
group & behavioral acc & $\C$-decode (correct key) & $\C$-decode (wrong key) \\
\midrule
failed baselines ($n{=}13$) & ${\sim}3$--$14\%$ & \textbf{52.3\%} & 5.2\% \\
failed \ns\ ($n{=}7$)       & ${\sim}3\%$       & \textbf{52.3\%} & 4.8\% \\
solved baselines ($n{=}3$)  & 93.3\%            & 88.4\%          & 5.0\% \\
solved \ns\ ($n{=}9$)       & 97.1\%            & 98.3\%          & 5.0\% \\
\bottomrule
\end{tabular}
\caption{Ridge decode of the raw memory state (32 checkpoints from the
replication sweep; features $\kk_u^{\!\top}\C_t$ at the pair's write position;
held-out sequences). Failed models retain substantial linearly decodable information despite low
behavioral accuracy.}
\label{tab:decode}
\end{table}

\textbf{Failed models retain substantial decodable information}: models at
chance carry ${\sim}52\%$ linearly decodable, key-specific associations they
cannot behaviorally retrieve (Table~\ref{tab:decode}). The plateau is a
readout-learning failure over partially intact storage; escape co-develops both
(and \ns-trained solutions end with cleaner storage than baseline ones, 98\%
vs.\ 88\% decodable). This dissociation is consistent with the limited benefit of
delta-rule writes. Two controls sharpen the picture: the decode succeeds only
with the key from the pair's value position (where the causal convolution
has folded key and value together before the write); and a naive oracle-query
substitution ($\q_t:=\kk_u$ inside the full forward) fails its positive control
(solved models drop from 93--97\% to 19--40\%), indicating that successful retrieval
depends on the learned query geometry.

\section{Practical guidance}
\label{sec:guidance}

\begin{itemize}
\item \textbf{Evaluate schedule changes first.} On plateau-escape tasks,
  deleting the LR decay yielded 8/16 solved seeds, compared with 9/16 for \ns, at the original
  step budget, at $1/6$ the per-step cost; per unit wall-clock the
  schedule changes outperform the orthogonalized read at this hardness
  (Fig.~\ref{fig:compute}). Cosine-to-budget couples the schedule to the
  budget and can suppress late escape.
\item \textbf{Differentiate through the whitened read and anneal it after escape.}
  Straight-through variants solved 0/16 seeds across the two tested configurations.
  Train with $\mathrm{NS}_k$ or
  an RLS read, anneal to the raw read late in training (gate the anneal on
  escape), and deploy a stock mLSTM. \textsc{rls} obtains the same benefit in a
  linear-cost recurrent form if per-position NS is too expensive.
\item \textbf{Audit recall-benchmark claims for censoring.} Solved-rate at one
  budget and schedule is sensitive to escape dynamics. Report escape-time
  survival curves, vary the schedule, and separate destabilization from
  plateauing as competing risks. The diagnostic probes here (state-spectrum
  traces, read-path gradient coherence) are cheap and trajectory-neutral.
\end{itemize}

\section{Discussion}
\label{sec:discussion}

\paragraph{Benchmark-driven architecture selection.} MAD and Zoology were built to
compare candidate architectures cheaply at small scale, on the premise that
small-scale rankings predict behavior at scale \citep{poli2024mad,arora2023zoology}.
Our results complicate that premise. On this benchmark, solved-rate at a fixed
budget is a censored measurement of escape hazard (\S\ref{sec:schedule}), and an
apparent architectural advantage can be produced by the learning-rate schedule
alone; in the case studied here the schedule accounts for most of the published
effect. To the extent that other recall benchmarks share the plateau-escape
training dynamics, a ranking taken at a single budget and schedule confounds
capability with small-scale trainability. Trainability is also the property most
sensitive to the changes that come with scale:
\S\ref{sec:basin}--\ref{sec:batch} show the workable-hyperparameter region moving
with learning rate, hardness, and batch size, so an advantage observed at 78K parameters
and 2{,}000 steps requires
validation under the intended production training configuration. The checks used in this paper are
inexpensive and could be adopted as standard practice: escape-time survival curves
in place of solved-rate at one budget, replication of the ranking under a second
schedule and batch size, separate accounting for destabilization as a competing risk,
and, where checkpoints allow, a probe for stored-but-unread
structure (\S\ref{sec:decode}).

\paragraph{Relation to emergent-ability claims.} The same measurement problem
appears, at much larger scale, in the debate over emergent abilities.
\citet{wei2022emergent} describe abilities that appear abruptly as compute grows;
\citet{schaeffer2023mirage} respond that many such thresholds are produced by
discontinuous metrics applied to smoothly improving models. That debate has mostly
been conducted through external scaling curves, because the internal state of the
models in question is difficult to instrument. The system studied here is small
enough to instrument completely, and it realizes the scenario of
\citet{schaeffer2023mirage} in a directly checkable form: the behavioral metric
sits at chance for thousands of steps while roughly half of the task-relevant
associations are already linearly decodable from the memory state
(\S\ref{sec:decode}), and escape is associated with learning in the readout pathway
(\S\ref{sec:mechanism}). These findings concern this task and training regime.
Plateau-escape recall tasks allow proposed explanations of abrupt behavioral
transitions to be tested against measurements of internal state.

\paragraph{Scaffolded training.} The procedure in
\S\ref{sec:scaffold} (optimize through a better-conditioned reparameterization
of the read, then anneal the reparameterization away before deployment) is related to
other training-time modifications: quantization-aware training inserts the quantizer
during
training and removes it afterwards \citep{jacob2018qat}, and batch-norm folding
and auxiliary losses likewise discard training-time structure at inference. Two
observations from this study seem relevant to that broader family. First, the
whitening helps only when the forward and backward computations agree
(\S\ref{sec:mechanism}), which suggests self-consistency as a requirement for
scaffolds of this type. Second, the connection to optimizer design is close: Muon
applies the same Newton--Schulz orthogonalization to gradient updates
\citep{jordan2024muon,wang2025muonmemory}, where it also improves conditioning
without changing the model class. Whether other conditioning problems admit
removable scaffolds of this kind is a question we leave open; this study
contributes one worked example.

\section{Related work}
\label{sec:related}

\paragraph{Recall as the discriminating capability.} Zoology and Based
\citep{arora2023zoology,arora2024based} identified precise recall as where the
attention-vs-recurrence gap concentrates; MAD \citep{poli2024mad} systematized
synthetic proxies. Write-rule design (delta-rule fast weights
\citep{schlag2021linear,yang2024deltanet}, gated matrix memories
\citep{beck2024xlstm}, selective state spaces \citep{gu2023mamba}) is largely
contention management for a fixed-size store. \citet{wang2025testtime} formalize
this design space as test-time regression, with linear attention, the delta rule,
and recursive least squares arising as choices of regressor and optimizer; our
elimination table is an empirical probe of that taxonomy under training dynamics.
The taxonomy distinguishes delta's online gradient descent from RLS's exact
solve, which produce different outcomes in our experiments. Across the tested
variants, improved trainability is associated with self-consistent conditioned
reads (NS, RLS); raw, normalization-only, and straight-through reads have lower
solved fractions. These results also inform benchmark use
(\S\ref{sec:discussion}): at the tested scale, optimization dynamics substantially
affect MAD noisy-recall solved-rates and can contribute to gains attributed to a
write rule or read modification. \citet{schaeffer2023mirage} make the parallel argument
about emergent-ability claims from discontinuous metrics; our storage/readout
dissociation (\S\ref{sec:decode}) supplies the internal-state version.

\paragraph{Plateau-escape dynamics.} The chance-plateau / sharp-escape shape,
its learning-rate and schedule sensitivity, and the role of gradient noise are
shared with the grokking literature \citep{power2022grokking}. The settings
differ in kind (grokking is delayed generalization after train-set
memorization, while query accuracy in our task remains near chance during the plateau). Escape
analysis is applicable to both settings, and \citet{ersoy2026metastable} derive a noise-driven
escape law in the same $(\eta, B)$ variables our \S\ref{sec:batch} fit measures
empirically.

\paragraph{Orthogonalization in learning systems.} Newton--Schulz orthogonalization
\citep{bjorck1971iterative,higham2008functions} has recently been prominent in
optimizer design: Muon \citep{jordan2024muon,bernstein2024old} orthogonalizes
gradient updates, and \citet{wang2025muonmemory} explains its advantage over Adam
via more isotropic singular spectra in associative-memory parameters, a
related conditioning effect in parameter space. POGO
\citep{javaloy2026pogo} maintains orthogonal parameter matrices at scale; in our
tests parameter orthogonality (pogo\_qk) does not reproduce the state-whitening
benefit, supporting a role for conditioning the evolving memory state. On the
architecture side, the Mesa layer and MesaNet
\citep{vonoswald2023mesa,vonoswald2025mesanet} deploy the recursive least-squares
read our \textsc{rls} variant implements, and Variational Linear Attention
\citep{pandey2026vla} builds a Sherman--Morrison regularized memory with recall
gains attributed to interference control; our swap and anneal results suggest a
testable reinterpretation of such designs at small scale: conditioned reads of
this family may improve trainability and permit removal through annealing
after escape.

\paragraph{Emergence and metric artifacts.} \citet{wei2022emergent} catalogued
sharp capability thresholds in large models; \citet{schaeffer2023mirage} argued many
are artifacts of discontinuous metrics over smoothly improving competence. Our
\S\ref{sec:decode} is a mechanistic, internally-instrumented instance of that
argument on a plateau-escape recall task (\S\ref{sec:discussion}).

\paragraph{The proposal under study.} \citet{tambde2026ortho} motivates read-time
orthogonalization by recall needs in attention-free settings such as Dreamer-style
world models \citep{hafner2020dreamer}. Our replication confirms the observation;
our mechanism experiments support an optimization explanation and demonstrate
that the intervention can be removed after training.

\section{Limitations}
\label{sec:limitations}

This is a mechanism study of one intervention on one synthetic task family and one
small architecture ($\sim$78K-parameter one-block mLSTM), in the regime where the
original claim was made. The training-scaffold interpretation is supported in this
setting; whether it
transfers to multi-block models, other recall families, or language-model
scale is untested. Several comparisons use 8--16 seeds per cell; we quantify uncertainty with
survival analysis and exact tests, and we release every per-seed trace. The
hazard model treats destabilization as independent censoring; a joint
competing-risks model might refine the analysis of the high-learning-rate regime. Two
secondary negatives are
reported as such (out-of-sample failure of probe-based seed triage; the
init$\times$data lottery being dominated by their interaction), and the
\textsc{pogo}+\ns\ stacking cell remains optimizer-confounded.

\section*{Reproducibility}

All code (task generator vendored, models, training loop, probes, analysis, and
figure scripts), per-seed results with full evaluation traces, and probe telemetry
are released at \url{https://github.com/no-way-labs/recurrent-memory-scaffold}.
Every number in this paper regenerates from \texttt{runs/*.jsonl} via the analysis
scripts. Experiments ran mostly on NVIDIA L4s (Modal, $\le$10 concurrent), with a
single vocab-96 batch-64 arm on an A100; total compute cost of the study was
approximately \$120. Probes are
trajectory-neutral: training with probes enabled is bit-identical to training
without.

\bibliographystyle{abbrvnat}
\bibliography{refs}

\end{document}